\documentclass[sigconf]{acmart}
\renewcommand\footnotetextcopyrightpermission[1]{} 
\makeatletter
\renewcommand\@formatdoi[1]{\ignorespaces}
\makeatother

\usepackage{amsmath}

\AtBeginDocument{%
  \providecommand\BibTeX{{%
    \normalfont B\kern-0.5em{\scshape i\kern-0.25em b}\kern-0.8em\TeX}}}

\begin{document}
\pagestyle{plain}

\title{Learning Discretized Bayesian Networks with GOMEA}

\author{Damy M.F. Ha}
\email{D.M.F.Ha@lumc.nl}
\affiliation{%
  \institution{Leiden University Medical Center}
  \city{Leiden}
  \country{The Netherlands}}

\author{Tanja Alderliesten}
\email{T.Alderliesten@lumc.nl}
\affiliation{%
  \institution{Leiden University Medical Center}
  \city{Leiden}
  \country{The Netherlands}}

\author{Peter A.N. Bosman}
\email{Peter.Bosman@cwi.nl}
\affiliation{%
  \institution{Centrum Wiskunde \& Informatica}
  \city{Amsterdam}
  \country{The Netherlands}}

\renewcommand{\shortauthors}{Ha, et al.}

\begin{abstract}
Bayesian networks model relationships between random variables under uncertainty and can be used to predict the likelihood of events and outcomes while incorporating observed evidence. From an eXplainable AI (XAI) perspective, such models are interesting as they tend to be compact. Moreover, captured relations can be directly inspected by domain experts. In practice, data is often real-valued. Unless assumptions of normality can be made, discretization is often required. The optimal discretization, however, depends on the relations modelled between the variables. This complicates learning Bayesian networks from data. For this reason, most literature focuses on learning conditional dependencies between sets of variables, called structure learning. In this work, we extend an existing state-of-the-art structure learning approach based on the Gene-pool Optimal Mixing Evolutionary Algorithm (GOMEA) to jointly learn variable discretizations. The proposed Discretized Bayesian Network GOMEA (DBN-GOMEA) obtains similar or better results than the current state-of-the-art when tasked to retrieve randomly generated ground-truth networks. Moreover, leveraging a key strength of evolutionary algorithms, we can straightforwardly perform DBN learning multi-objectively. We show how this enables incorporating expert knowledge in a uniquely insightful fashion, finding multiple DBNs that trade-off complexity, accuracy, and the difference with a pre-determined expert network.

\end{abstract}




\maketitle

\section{Introduction}
Bayesian Networks (BNs) \cite{koller_friedman_probabilistic_graphical_models, pearl_probabilistic_reasoning} are probabilistic graphical models that model relationships between random variables under uncertainty. The relationships between variables can be depicted using a Directed Acyclic Graph (DAG). The process of optimizing the DAG for given (tabular) data, which is often called structure learning, has been extensively researched in the literature (e.g., \cite{koller_friedman_probabilistic_graphical_models, bn_gomea}) and applied to many real world applications such as in the medical domain: \cite{example_radiologist, example_gene_expression, example_endorisk}, geology and environmental modeling domain: \cite{example_bit_rate, example_coastal_erosion, example_discretizing_environmental_data}, and (risk and safety) management: \cite{example_chemical_process, example_supply_chain, example_aviation}.

In the aforementioned domains, it is not uncommon to have a mix of discrete and continuous random variables. How to best incorporate continuous variables is however not straightforward. In the literature, there are various methods to extend discrete BNs with continuous variables. For example, a common method is to call upon a domain expert, who is tasked to pre-discretize continuous variables before structure learning or to model the continuous variables with a parametric distribution. It might however be difficult to consult a domain expert or they might not always be able to correctly model the variables. Non-parametric modelling of variables \cite{mixbn, nonparametric_bayesian_networks} on the other hand, does not require expert knowledge. However, in non-parametric models, normality is usually assumed. Discretization techniques \cite{fayyad_irani, discretizing_friedman, learn_dbn, example_bit_rate, joes_cousin} offer an alternative as neither expert knowledge is required a priori, nor must the assumption of normality hold. The optimal discretization however, depends on the relations modelled between the variables, necessitating simultaneous optimization.

In this work, for the first time, a state-of-the-art structure learning approach based on the Gene-pool Optimal Mixing Evolutionary Algorithm (GOMEA) family of algorithms \cite{bn_gomea} is extended to jointly learn variable discretizations. The proposed Discretized Bayesian Network GOMEA (DBN-GOMEA) is compared to the state-of-the-art on randomly generated problems. When the algorithms are tasked to retrieve randomly generated ground-truth networks, it is shown that DBN-GOMEA obtains, similar or better performance than the state-of-the-art. Moreover, leveraging key strengths of EAs in multi-objective optimization, it is possible to straightforwardly perform DBN learning multi-objectively. The proposed approach is fundamentally different from e.g., \cite{mo_bi_objective_scalable}, where a bi-objective search is performed on (proxies of) the accuracy and complexity and e.g., \cite{exploiting_expert_knowledge}, where prior knowledge is included in the search by altering prior model probabilities according to expert knowledge. Our multi-objective approach leverages a tri-objective search to incorporate expert knowledge in a uniquely insightful fashion that enables finding multiple discretized BNs that trade-off (proxies of) the model accuracy, complexity, and difference to a pre-determined expert network.

The code is available at: \url{https://github.com/damyha/dbn_gomea}.

\section{Discrete Bayesian Networks}
\label{s:discrete_bayesian_networks}
\begin{figure}[h]
  \centering
  \includegraphics[width=0.9\linewidth]{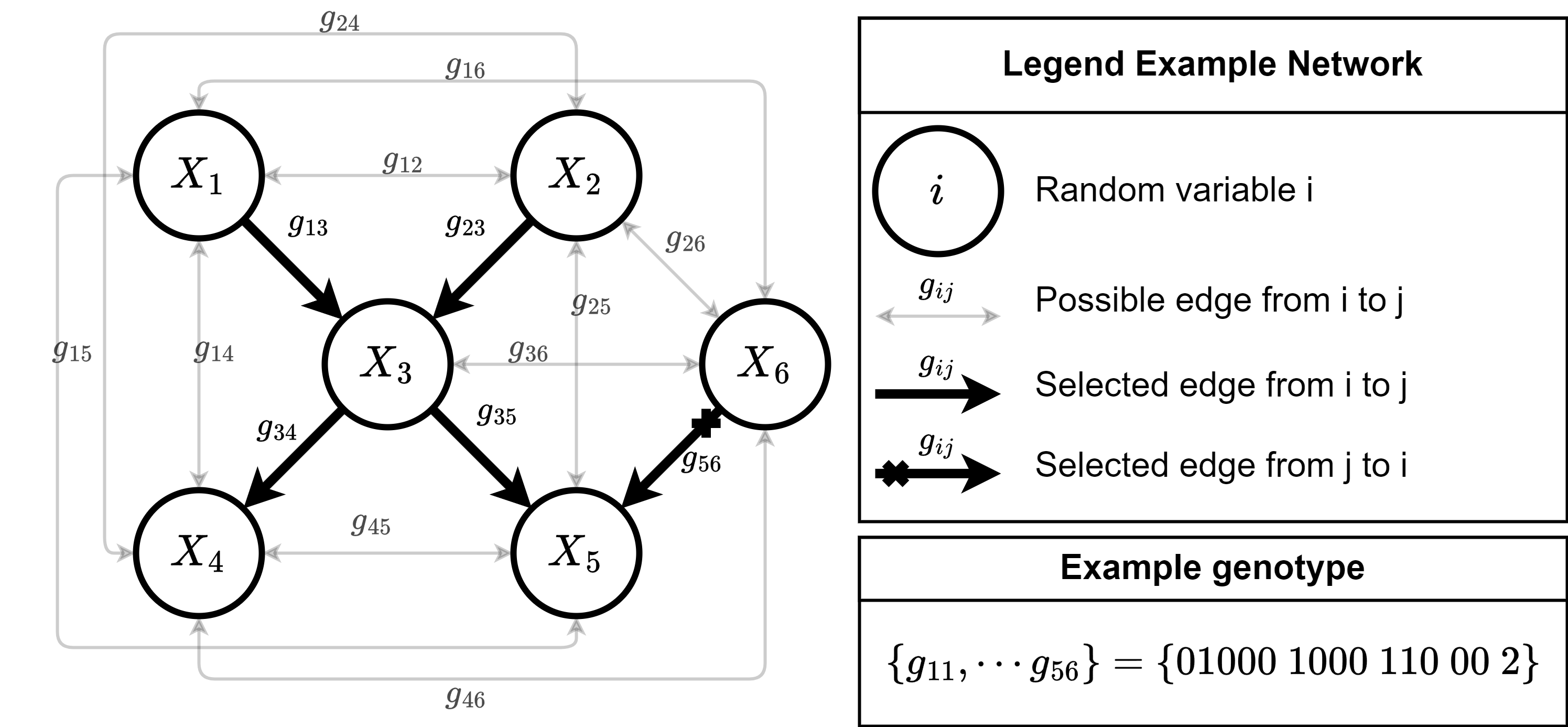}
  \caption{Example of a DAG used to represent a BN (in black) and all possible edges (in grey).}
  \label{fig:example_bn}
\end{figure}

BNs \cite{koller_friedman_probabilistic_graphical_models, pearl_probabilistic_reasoning} are a class of probabilistic graphical models. A BN $B$ is defined by a DAG $G$, which represents $ \textbf{X} = \left\{ X_1, \cdots X_N \right\}$ random variables. Each node $i$ in $G$ is associated with a random variable $X_i$ and has a (conditional) probability distribution $P(X_i|\text{pa}(X_i))$, where the probability of $X_i$ is conditionally dependent on the parent nodes of $X_i$, i.e., $\text{pa}(X_i)$. In $G$, this relationship is modeled via a directed edge from each of the parent nodes to node $i$. An example of $G$ is shown in Figure \ref{fig:example_bn}, where $\text{pa}(X_3) = \left\{ X_1, X_2 \right\}$, and $X_3$ is a parent of $X_4$. Node $X_3$, together with spouse $X_6$ are also parents of $X_5$. Given $G$ and all conditional probabilities $\Theta$, the probability of $\textbf{X}$ can be written as a product of the individual conditional node probabilities, as is shown in Equation \ref{eq:p_bayesian_network}.

\begin{equation}
\label{eq:p_bayesian_network}
    \text{P}(X_1, \cdots, X_{N}) = \prod_{i=1}^{N} \text{P}(X_i|\text{pa}(X_i))
\end{equation}

\subsection{Bayesian Network GOMEA}
\label{ss:bn_gomea}
In recent work, a state-of-the-art score-based BN structure learning algorithm was developed, called BN-GOMEA \cite{bn_gomea}. BN-GOMEA employs an Evolutionary Algorithm (EA) from the Gene-pool Optimal Mixing Evolutionary Algorithm (GOMEA) family. BN-GOMEA learns network structures from discrete data. It showed superior performance to other EAs, greedy hill-climbing, and tabu-list based algorithms such as Ordering-Based Search, Sparse Candidate, and Max-Min Hill-Climbing.

In BN-GOMEA, the BN structure learning problem is formulated as follows: solutions are represented as a string of discrete variables. Each variable in the string represents an edge between an arbitrary random variable $i$ and random variable $j$ (where $i \neq j$). The value of each variable can be 0, meaning no edge between $i$ and $j$, 1, meaning a directed edge from $i$ to $j$ or 2, meaning a directed edge from $j$ to $i$. The problem formulation results in $l_{\text{total}} = \frac{N}{2}(N-1)$ number of variables to represent all edges in the graph, where $N$ is the total number of random variables. An example of a BN with a genotype representation is given in Figure \ref{fig:example_bn}. This problem formulation however, allows cyclic networks to exist. Therefore, a repair operator is used to remove cycles. To check if solutions contains a cycles, a depth-first search is executed. If a cycle is found, the last edge that completes the cycle in the depth-first search is removed.

BN-GOMEA, makes use of a linkage model that captures interdependencies between problem variables. A linkage model is made up of Family of Subset (FOS) elements, where each FOS element is a set of indices that indicate dependencies between the problem variables represented by those indices. The FOS elements are used during variation to effectively mix groups of variables to create fitter solutions. In \cite{bn_gomea}, BN-GOMEA makes exclusively use of the linkage tree. The linkage tree is hierarchical tree structure which is learnt from the population. The Gene-pool Optimal Mixing (GOM) variation operator leverages this linkage tree. Each solution in the population undergoes GOM. First, the solution is cloned. Then the linkage tree is randomly traversed for this solution. For each FOS element in the linkage tree, a random donor solution is selected from the population. The variables, as indicated by the FOS element, are then copied to the offspring solution. If the change results in a worse fitness, the change is reverted in the offspring, otherwise the change is kept.  

Other than the GOM operator, the excellent performance of BN-GOMEA can also be attributed to two other reasons. First, BN-GOMEA exploits the use of partial evaluations in combination with the linkage tree. When an offspring solution is created from a parent solution, each GOM step only changes part of the solution. It is more efficient to only recalculate the fitness contribution of variables that have changed, if the fitness function is decomposable. For typical fitness functions used with BNs, this is the case. In BN-GOMEA the decomposable BDeu score was used.

The second reason for BN-GOMEA's excellent performance is because a local search operator is additionally used. Upon initialization and after applying GOM to every solution in the population, the local search operator is applied on every solution in the population. The local search operator randomly traverses all variables of a solution and evaluates the fitness when the selected edge takes a different value, i.e., any value in $\left \{ 0, 1, 2 \right \}$ different from the current value. During local search, only changes that result in a better fitness are accepted, otherwise the change is reverted.

At last, BN-GOMEA makes use of an Interleaved Multi-start Scheme (IMS), which runs multiple populations of various sizes side by side. The IMS avoids the user to excessively tune the population size manually. For this, the IMS ensures that a population of size $n_{\text{pop}}$, executes 4 generations before a population of $2 \cdot n_{\text{pop}}$ executes a single generation, starting from a base population of size 2.

\section{Discretization of Continuous Random Variables in Bayesian Networks} 
\subsection{DBN-GOMEA}
In this work, BN-GOMEA is extended such that it can handle continuous random variables without prior discretization, i.e., the variables are discretized during structure learning. This new algorithm is dubbed Discretized Bayesian Network-GOMEA (DBN-GOMEA).

First, the BDeu score used in BN-GOMEA is replaced by a density based function, as the variables are reinterpreted in terms of density. The assumption that is made is that if data is discretized, it is uniformly distributed within that discretization. To optimize the uniform discretizations, the density  of the discretizations should be maximized. As such, the log likelihood over the densities is taken as fitness function, as is shown in equation \ref{eq:fitness_density}, where $\mathbf{x_i} \in \mathbb{R}^N$ is training sample $i$ from a training data set and $n$ the size of the training data set. To make sure that the density is invariant to the range of data, the data ranges are normalized to $\left [ 0, 1 \right ]$ prior to calculating the densities. This is similar to what has been done in \cite{joes_cousin}. As a penalty term, the penalty of the BIC score \cite{bic} is used, where the model complexity $\text{C}(G)$ is dependent on the number of parent discretizations: $ \left| \text{pa}(X_i) \right|$, the number of discretizations of $X_i$: $ \left| X_i \right|$, and $n$ as shown in Equation \ref{eq:fitness_penalty}. This results in the fitness function, displayed in Equation \ref{eq:fitness}.

\begin{equation}
    \label{eq:fitness_density}
    \text{LL}(\textbf{X}, G) = \prod_{i=1}^{n} \log(f_{\text{density}}(\mathbf{x_i}))
\end{equation}

\begin{equation}
    \label{eq:fitness_penalty}
    \text{C}(G) = \sum_{i=1}^{N} \left| \text{pa}(X_i) \right| \cdot (\left| X_i \right| - 1) \cdot \log(\frac{n}{2})
\end{equation}

\begin{equation}
    \label{eq:fitness}
    \text{fitness}(\textbf{X}, G) = \text{LL}(\textbf{X}, G) - \text{C}(G)
\end{equation}

To discretize continuous variables, two common discretization methods are introduced, namely: Equal-Width (EW) and Equal-Frequency (EF). In EW discretization, data is split into 'k' equally ranged discretization bins. In EF discretization, data is sorted and split into 'k' equally filled discretization bins. In DBN-GOMEA, the number of discretizations 'k' is optimized by appending the discretization counts 'k' of each continuous random variable to the solution representation of BN-GOMEA, i.e., the representation is enlarged with $N_c$ variables where $N_c$ is the number of continuous variables.

As the solution representation is altered, the local search operator of BN-GOMEA is extended. In DBN-GOMEA, the original local search operator for the network topology is kept. However, when a solution variable is selected that represents a discretization count, the modified local search operator increases and decrease the discretization count by one, i.e., $\left\{ k - 1, k + 1 \right\}$. If the resulting number of discretizations falls outside the minimum or maximum number of discretizations, which are 2 and 15 respectively, the local search step is not executed. In this work, the minimum and maximum discretizations have been chosen to keep computation times feasible.

\subsection{Post-structure Learning Discretization}
Although EW and EF discretization are commonly used in the literature, in practice it is unlikely for data to be EW or EF distributed. As a consequence, an inaccurate discretization might be found. For this reason, the effect of optimizing the discretization boundaries will be investigated. This will however be done after structure learning has finished, as structure learning and discretization can become expensive when the sample size grows. 

The algorithm selected for this task is the Real-Valued GOMEA (RV-GOMEA) \cite{rv_gomea, gomea_library}, which is a state-of-the-art real-valued optimization algorithm. With the network structure and number of discretizations for each continuous random variable fixed, the boundaries can be optimized using the same density fitness function. As a result, the log likelihood over the densities is potentially further optimized, without a change in complexity.

The boundary optimization in RV-GOMEA is encoded by concatenating all boundaries to be optimized into a single solution. Instead of directly optimizing the boundaries of the data, the optimization problem is reformulated by sorting the unique data values $\textbf{u}$ of each continuous random variable and to optimize the sample indices that separate the data. By optimizing the sample indices, the flat-landscape between samples becomes equiprobable, compared to directly optimizing the boundaries. The boundary at sample index $i$ is then calculated by taking the midway point between sample $u_i$ and sample $u_{i+1}$. As RV-GOMEA optimizes real-valued problems, the solution parameters (which represent the sample indices) are rounded down.

For RV-GOMEA, the linkage tree is once again used as a linkage model. 

\subsection{Bayesian Method}
In \cite{learn_dbn}, a discretization method is proposed that finds a discretization $\Lambda$ that maximizes a likelihood score, given a BN structure. The method uses Bayes rule to maximize: $P(\Lambda) \cdot P(D | \Lambda)$, where $P(\Lambda)$ is the prior of a discretization policy and $P(D | \Lambda)$ the probability of the data given the discretization policy. The likelihood is formulated by making assumptions, of which one assumption is that the prior probability of a discretization boundary between two unique consecutive sample values is proportional to their difference. 

Maximization of the likelihood is implemented via dynamic programming. By doing pre-calculations, the discretization runtime is reduced to $\mathcal{O}(r \cdot n^2)$, where $r$ is a constant and $n$ the sample size. 

In \cite{learn_dbn} furthermore, this Bayesian discretization method is combined with a structure learning algorithm. The combination of both algorithms is dubbed LDBN in this work. In LDBN, the structure is learnt by first applying EW discretization on all continuous random variables, where the number of discretizations is selected to be the largest number of instantiations amongst all discrete random variables. After EW discretization, an arbitary random variable is selected as a starting point. The remaining random variables are randomly selected and sequentially added to the BN structure as child nodes. An edge between the new child node and potential parent node(s) materializes when the K2 score of the network improves after adding the edge. If a new edge is added to the network, the Bayesian discretization method is applied on all nodes that fall within the Markov blanket of the new node. All nodes within the Markov blanket are sequentially discretized in a random order. 

As both the structure learning algorithm as well as the Bayesian discretization contain randomness, LDBN runs the structure learning and discretization algorithm multiple times. We kindly refer to \cite{learn_dbn} for more details.

\section{Multi-Objective Learning}
\label{s:mo_bayesian_networks}
A major limitation of Single-Objective (SO) BN learning is that the weight of the complexity term is not straightforward to set \cite{mo_bi_objective_scalable}. Furthermore, the obtained model might not be trusted by an expert when the expert has their own beliefs. Taking a Multi-Objective (MO) perspective can offer a solution to these problems. First, using MO search does not require the user to know the penalty factor a priori. Furthermore, the search returns many networks out of which a domain expert can choose a network that matches (partly) with their own prior belief or discover new knowledge this way.

\begin{figure}[h]
  \centering
  \includegraphics[width=1.0\linewidth]{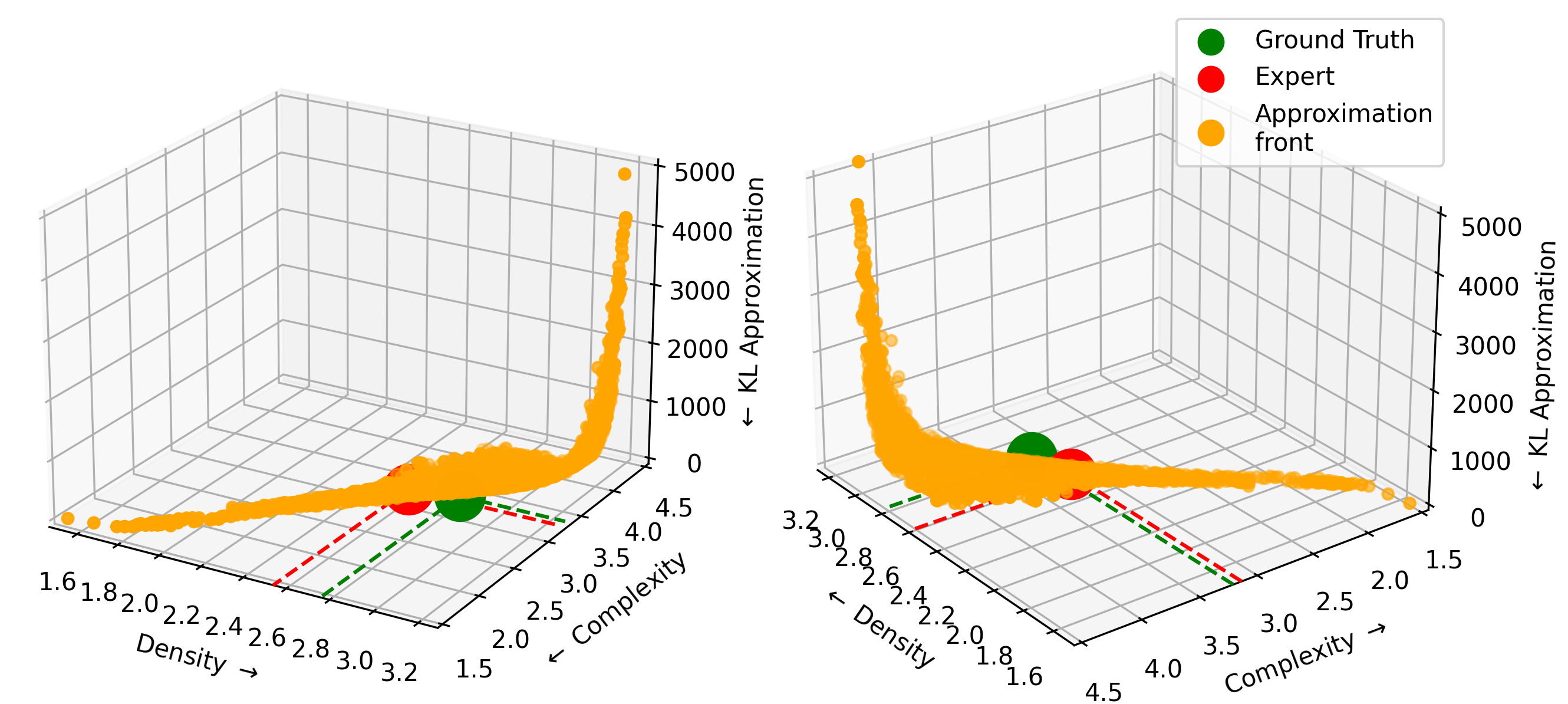}
  \caption{Impression (rotated visualizations) of the approximation front of a multi-objective run together with the objective values of the ground truth and expert solutions. The x, y, and z axis are on a log-log-linear scale. }
  \label{fig:example_mo}
\end{figure}

A straightforward way to do MO search is to optimize (a proxy of) accuracy and model complexity as in e.g., \cite{mo_bi_objective_scalable}, where an EA is used, performing MO search relatively straightforward. For GOMEA, multi-objective variants also exist that are direct extensions of the SO versions, necessitating no further adaptions to e.g., DBN-GOMEA's genotype or operator. Using the density function as is, an MO version of the problem can be created straightforwardly by making Equation \ref{eq:fitness_density} and Equation \ref{eq:fitness_penalty} separate objectives. See Section \ref{ss:mo_dbn_gomea} for more on this. A subset of the networks obtained from this MO search can then be shown to an expert, who decides which network is most appropriate, observing the fit to the data and matching with their own beliefs. The inclusion of expert knowledge however, could provide additional guidance to the search. In this work, not only are the density and complexity optimized as separate objectives, the difference with an a priori determined expert network is also optimized. To this end, the Kullback-Leibler (KL) divergence is used as a distance between a candidate network and an expert network. The KL divergence is shown in Equation \ref{eq:kl_approx}, where $P(\mathbf{X})$ is the probability distribution of the expert network, $Q(\mathbf{X})$ the probability distribution of a candidate network, and $\mathcal{X}$ the sample space. The KL divergence is 0 when two probability distributions are identical, and is larger than 0 otherwise. An example of the objective space of a resulting MO run is shown in Figure \ref{fig:example_mo}.

\begin{equation}
    \label{eq:kl_approx}
    D_{KL}(P || Q) = \sum_{x \in \mathcal{X}} P(\mathbf{X}) \log\left ( \frac{P(\mathbf{X})}{Q(\mathbf{X})} \right )
\end{equation}

\subsection{MO-DBN-GOMEA}
\label{ss:mo_dbn_gomea}
For the MO optimization, Multi-Objective Gene-pool Optimal mixing Evolutionary Algorithm (MO-GOMEA) \cite{mo_gomea} is used. MO-GOMEA is a state-of-the-art multi-objective EA that is also part of the GOMEA family. It can similarly exploit partial evaluations for enhanced efficiency. MO-GOMEA uses domination-based optimization, i.e., it uses the concept of Pareto dominance to find better solutions. A solution is said to Pareto dominate another solution if it is not worse in any objective and better in at least one objective.

Given $m$ objectives that need to be optimized, a population in MO-GOMEA is partitioned into $c$ clusters of equal sizes. For each cluster, a linkage model is learnt. In this work, the linkage tree is used, which is similar to the linkage tree in Section \ref{ss:bn_gomea}. Each cluster is evolved using the respective linkage model. A select number of clusters (specifically $m$) with the (respective) highest average objective values are selected to optimize the individual objective functions in a SO setting using the SO GOM operator. For the remaining clusters the MO GOM operator is used. Different from the SO GOM operator, the MO GOM operator accepts a solution when any of the following holds: 1) the GOM altered solution dominates the unaltered solution, 2) the altered solution has the same objective values, 3) the altered solution is not dominated by any solution in the elitist archive. The elitist archive is an archive of non-dominated solutions found during the optimization. In this work, an elitist archive size of 10,000 is used to collect as many solutions as possible during the search, while balancing computation time.

Similar to BN-GOMEA, MO-GOMEA uses the IMS to manage its population size. However, the IMS in MO-GOMEA additionally manages the number of clusters $c$ in a population. The population size starts at 8 and is multiplied by 2 for every new population size. The number of clusters starts at $m$ + 1, and is incremented by 1 for every new population. For more details, we kindly refer to \cite{mo_gomea}.

In this work, MO-GOMEA is slightly extended by making MO-GOMEA capable of solving discrete problems over binary problems only. For this, the suggestions proposed in \cite{mo_gomea} are followed. The most important change made, is replacing the binary linkage tree in MO-GOMEA with the discrete linkage tree of \cite{bn_gomea}. Furthermore, the Bayesian network structure learning, as proposed in Section \ref{s:discrete_bayesian_networks} is integrated into MO-GOMEA. The new structure learning algorithm is dubbed MO-DBN-GOMEA.

\section{Experiments and results}
\subsection{Network Generation}
In this work, randomly generated BN structures and probability distributions are used to assess the performance of the algorithms. For this, the network generator algorithm of \cite{random_bn_structure} is used to generate random BNs. Probability distributions are generated using a method described below. Data sets are sampled from the ground truth networks and given to the algorithms. In \cite{random_bn_structure}, random BN structures are generated under constraints. The maximum number of parent random variables are chosen to be 6 and the maximum number of edges in a network are set to be at most 40\% of all possible edges $l$. These constraints have been chosen, such that networks can be evaluated within reasonable time.

The probability distributions are randomly generated by first separating the random variables into discrete and continuous variables. The number of discrete variables is set to 10\% with a minimum of at least one discrete variable per ground truth network. Each random variable, whether discrete or continuous, is then randomly assigned between 2 and 6 discretizations, e.g., if a random variable is randomly assigned 5 discretizations, the possible values are: $\left\{ 1, 2, 3 , 4, 5 \right\}$. A discrete probability table is then generated for each random variable, that maps the possible parent values to a probability of a specific discretization value. In this work, the probability tables are generated in three ways: EW, EF or random probability distributions. For EF probability distributions, the probability of sampling any value is equiprobable. For EW and random probability distributions, random probability tables are generated.

Discrete samples can now be retrieved. For the continuous variables however, the discrete probability tables must be converted to continuous probability distributions. For this, a mapping is generated that maps each discrete value to a specific range of continuous values. Continuous samples can be obtained by uniformly sampling from this range. For example, if a continuous random variable has 3 discretizations with ranges: $\left [ 1.0, 2.0 \right \rangle$, $\left [ 2.0, 2.5 \right \rangle$, $\left [ 2.5, 3.5 \right \rangle$, and a discrete value of 2 is sampled, a continuous sample is produced by uniformly sampling from $\left [ 2.0, 2.5 \right \rangle$. The sample ranges are designed to be adjacent to each other and non-overlapping. For EW probability distributions, the sample ranges are set to be equally spaced. For EF and random probability distributions, the sample ranges are determined randomly. 

\subsection{Metrics}
The performance of the algorithms is assessed using various metrics. In terms of network structure metrics, two metrics are used. First, the accuracy is defined as the sum of the number of correctly identified edges $TP$ and correctly identified absent edges $TN$ of a candidate network structure as a percentage of the sum of the total number of edges $l_{\text{total}}$ of the ground truth network (see Equation \ref{eq:accuracy}). A correctly identified edge is defined as an edge in the ground truth network which also appears in the candidate network without regarding the directionality of the edge. Note that this definition is different from other works such as e.g., \cite{empirical_score_functions}. A correct absent edge is defined as an absent edge in both the ground truth network and candidate network. The sensitivity is defined as the number of correctly identified edges $TP$ as a percentage of the total number of edges in the ground truth network: $l_{\text{edges}}$ (see Equation \ref{eq:sensitivity}).

\begin{equation}
    \label{eq:accuracy}
    \text{Accuracy} = \frac{TP + TN}{l_{\text{total}}}
\end{equation}

\begin{equation}
    \label{eq:sensitivity}
    \text{Sensitivity} = \frac{TP}{l_{\text{edges}}}
\end{equation}

To assess the quality of the discretizations, the KL divergence with respect to the ground truth network is used. The KL divergence was already introduced in Section \ref{s:mo_bayesian_networks}.

\subsection{Single-Objective Scalability}
\subsubsection{Single-Objective Scalability in Terms of Sample Size}
\label{ss:exp_sample_size}
\begin{figure*}[h]
  \centering
  \includegraphics[width=\linewidth]{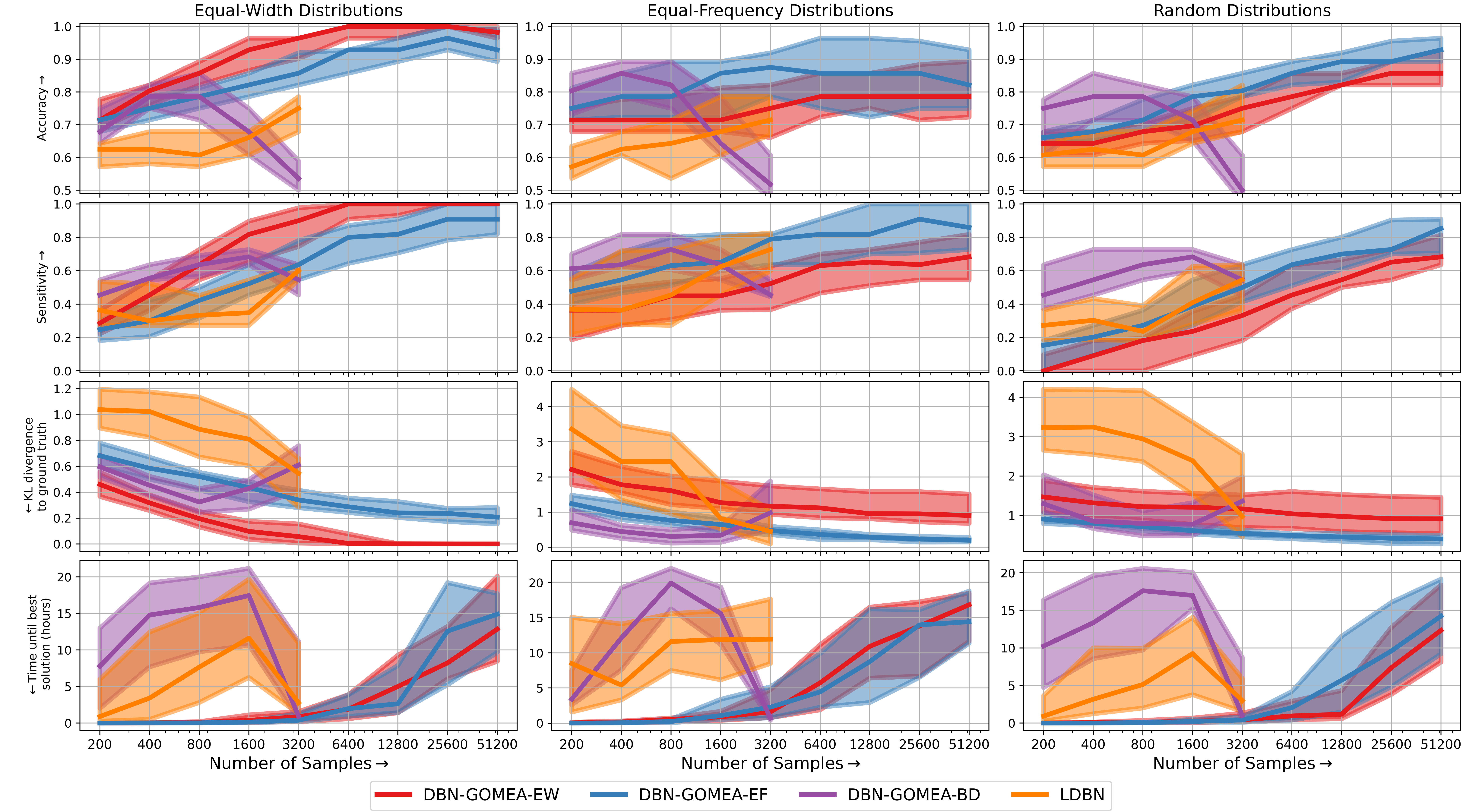}
  \caption{Scalability in terms of sample size for 30 random networks with 8 random variables having EW, EF and Random probability distributions. The solid lines are medians, while the shaded areas encompass the first and third interquartile ranges. The arrows on the y-axis point in the direction of improvement per metric.}
  \label{fig:exp_samples}
  \vspace*{-4mm}
\end{figure*}

\hfill\\
The scalability in terms of sample size is shown for various algorithms in Figure \ref{fig:exp_samples}. For this, 30 ground truth networks, each having 8 random variables, are generated with EW, EF and random probability distributions. To asses the KL divergence, 50.000 test samples are generated. DBN-GOMEA with EW, EF, and Bayesian Discretization (BD), as well as the structure learning algorithm from \cite{learn_dbn} (LDBN) are considered in these experiments. DBN-GOMEA-EF and DBN-GOMEA-EW are run on an Intel E5-2690 where each run uses a single core with 2GB of memory and 24 hours of computation time. The Bayesian discretization's memory requirements scale with $\mathcal{O}(n^2)$. Therefore, LDBN and DBN-GOMEA-BD are run on a (newer) E5-4650 with 20GB of memory per run and 24 hours of computation time. Due to computational constraints, it was not possible to run all algorithms on the E5-4650.

Figure \ref{fig:exp_samples} shows that for EW, EF, and random probability distributions, in general, DBN-GOMEA with the appropriate discretization techniques finds better network structures as well as better KL divergence when the sample size grows. Only DBN-GOMEA-EW obtains perfect network retrieval for EW problems, given at least 6400 samples. DBN-GOMEA-EF does not achieve perfect retrieval for EF problems, for any tested sample size. Note however, that the EF data is sampled from the ground truth network and thus not perfectly EF distributed. Hence; when using EF discretization, the boundaries are not optimal. Interestingly, LDBN in general, shows worse performance than DBN-GOMEA in terms of network metrics, KL divergence or in some cases in both. LDBN also runs out of memory when there are too many samples. DBN-GOMEA-BD also runs out of memory. For small sample sizes however, except for the EW problems, it can find better network structure and similar or better KL divergence compared to DBN-GOMEA-EW and DBN-GOMEA-EF. 

To test for differences in the KL divergence, a Mann–Whitney U statistical test is performed. Results obtained from ground truth networks with random probability distributions are investigated. In table \ref{tab:stat_test_sample}, the mean $\pm$ the standard deviation of the KL divergence is shown. Numbers in bold have the best average KL divergence or are statistically not different from the best. An alpha value of $0.05$ is used, with a Bonferroni correction of 63 as 27 tests are performed and 36 more tests will be performed later on. Table \ref{tab:stat_test_sample} shows that DBN-GOMEA-EF is the overall best, with DBN-GOMEA-BD performing similar on problems with few samples.

\begin{table}[]
\begin{tabular}{lcccc}
\hline
\begin{tabular}[c]{@{}c@{}}Number\\ of\\ samples\end{tabular} & \begin{tabular}[c]{@{}c@{}}DBN-\\ GOMEA-\\ EW\end{tabular} & \begin{tabular}[c]{@{}c@{}}DBN-\\ GOMEA-\\ EF\end{tabular} & \begin{tabular}[c]{@{}c@{}}DBN-\\ GOMEA-\\ BD\end{tabular} & LDBN \\ \hline
200 & 1.57 $\pm$  0.58 & \textbf{0.94} $\pm$  0.21 & 1.62 $\pm$  0.89 & 3.43 $\pm$  1.27 \\ 
400 & 1.45 $\pm$  0.59 & \textbf{0.83} $\pm$  0.19 & \textbf{1.16} $\pm$  0.78 & 3.39 $\pm$  1.17 \\ 
800 & 1.38 $\pm$  0.61 & \textbf{0.73} $\pm$  0.18 & \textbf{0.94} $\pm$  0.65 & 3.37 $\pm$  1.24 \\ 
1600 & 1.32 $\pm$  0.64 & \textbf{0.64} $\pm$  0.17 & \textbf{1.02} $\pm$  0.73 & 2.66 $\pm$  1.48 \\ 
3200 & 1.26 $\pm$  0.64 & \textbf{0.57} $\pm$  0.18 & 1.82 $\pm$  1.43 & - \\ 
6400 & 1.21 $\pm$  0.66 & \textbf{0.50} $\pm$  0.17 & - & - \\ 
12800 & 1.16 $\pm$  0.66 & \textbf{0.46} $\pm$  0.17 & - & - \\ 
25600 & 1.09 $\pm$  0.63 & \textbf{0.43} $\pm$  0.18 & - & - \\ 
51200 & 1.07 $\pm$  0.62 & \textbf{0.41} $\pm$  0.18 & - & - \\ 
\hline
\end{tabular}
\caption{The average KL divergence values and standard deviation to the ground truth networks of various algorithms and for different sample sizes. In bold are the best KL values and those statistically not different from it. The ground truth networks have random probability distributions.}
\label{tab:stat_test_sample}
\vspace*{-5mm}
\end{table}

\subsubsection{Single-Objective Scalability in Terms of Random Variables}
\label{ss:exp_nodes}
\begin{figure}[h]
  \vspace*{-3mm}
  \centering
  \includegraphics[width=\linewidth]{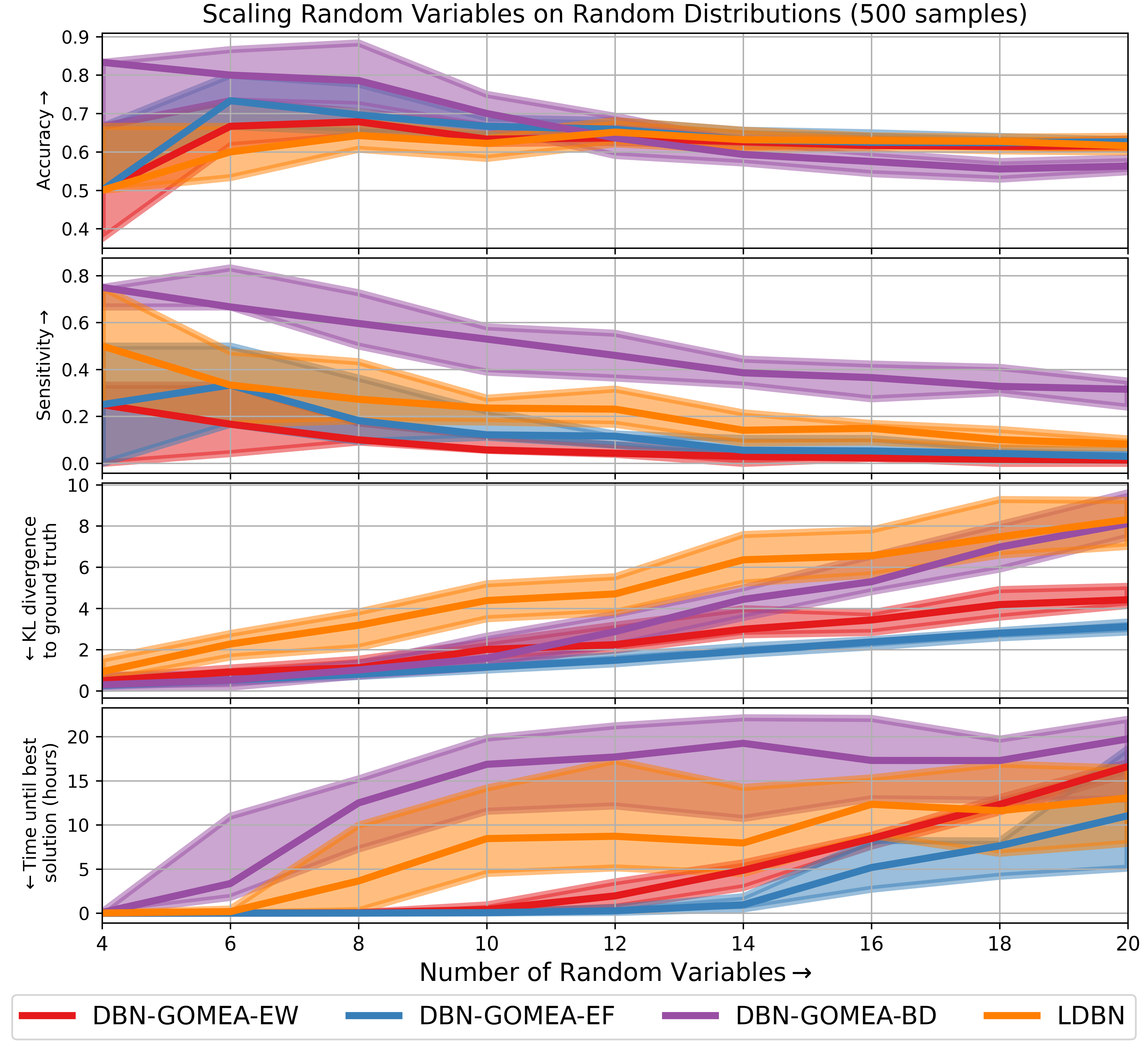}
  \caption{The scalability in terms of number of random variables. For each number of random variables on the x-axis, 30 ground truth networks were generated with random probability distributions. The solid lines are medians, while the shaded areas encompass the first and third interquartile ranges. The arrows on the y-axis point in the direction of improvement per metric.}
  \label{fig:exp_nodes_random}
  \vspace*{-3mm}
\end{figure}

\hfill\\
The scalability in terms of number of random variables in a network is shown in Figure \ref{fig:exp_nodes_random}. For this, 30 ground truth networks, with random probability distributions, are generated per ground truth network size. Each run was performed using 500 training samples, on a single core of an AMD Genoa 9654, with 2GB of memory and a computation budget of 24 hours.

Figure \ref{fig:exp_nodes_random} shows that DBN-GOMEA-BD obtains more accurate and more sensitive network structures when there are few nodes. However, after more than 12 nodes, the networks become less accurate compared to the ones obtained by the other algorithms. At the same time, the time it takes to find the best solution also nears the computation budget of 24 hours. This begs the question if DBN-GOMEA-BD needs more time to converge. The network accuracy obtained by the other algorithms seems to be similar, especially after 12 nodes.

Despite having similar network structures, the KL divergence does seem to differ per algorithm. To test for statistical differences in the KL divergence, a Mann–Whitney U statistical test is performed. In Table \ref{tab:stat_test_node}, the mean $\pm$ the standard deviation of the KL divergence is provided. Numbers in bold have the best average KL divergence or are statistically not different from the best. An alpha value of $0.05$ is used, with a Bonferroni correction of 63 as 27 tests were performed previously and 36 more tests are performed in Table \ref{tab:stat_test_node}.

Table \ref{tab:stat_test_node} shows that, similar to Table \ref{tab:stat_test_sample}, DBN-GOMEA-EF is amongst the best in terms of KL divergence. Table \ref{tab:stat_test_node} also shows that DBN-GOMEA-BD is amongst the best in terms of KL divergence when the number of random variables is small. LDBN and DBN-GOMEA-EW perform relatively poorly.

\begin{table}[]
\begin{tabular}{ccccc}
\hline
\begin{tabular}[c]{@{}c@{}}Random\\ Variables\end{tabular} & \begin{tabular}[c]{@{}c@{}}DBN-\\ GOMEA-\\ EW\end{tabular} & \begin{tabular}[c]{@{}c@{}}DBN-\\ GOMEA-\\ EF\end{tabular} & \begin{tabular}[c]{@{}c@{}}DBN-\\ GOMEA-\\ BD\end{tabular} & LBDN \\ \hline
4 & \textbf{0.58} $\pm$  0.45 & \textbf{0.29} $\pm$  0.14 & \textbf{0.45} $\pm$  0.40 & 1.11 $\pm$  0.81 \\ 
6 & 0.98 $\pm$  0.56 & \textbf{0.51} $\pm$  0.16 & \textbf{0.57} $\pm$  0.40 & 2.31 $\pm$  0.86 \\ 
8 & 1.26 $\pm$  0.51 & \textbf{0.78} $\pm$  0.18 & \textbf{1.09} $\pm$  0.63 & 3.04 $\pm$  1.05 \\ 
10 & 2.11 $\pm$  0.69 & \textbf{1.15} $\pm$  0.22 & \textbf{1.89} $\pm$  1.16 & 4.42 $\pm$  1.35 \\ 
12 & 2.57 $\pm$  0.91 & \textbf{1.54} $\pm$  0.25 & 3.02 $\pm$  1.06 & 4.63 $\pm$  1.07 \\ 
14 & 3.29 $\pm$  0.93 & \textbf{1.98} $\pm$  0.28 & 4.49 $\pm$  1.32 & 6.42 $\pm$  1.46 \\ 
16 & 3.47 $\pm$  0.82 & \textbf{2.32} $\pm$  0.29 & 5.69 $\pm$  1.25 & 6.97 $\pm$  1.73 \\ 
18 & 4.32 $\pm$  1.06 & \textbf{2.76} $\pm$  0.30 & 6.95 $\pm$  1.43 & 7.85 $\pm$  1.74 \\ 
20 & 4.66 $\pm$  0.82 & \textbf{3.14} $\pm$  0.33 & 8.33 $\pm$  1.77 & 8.35 $\pm$  1.52 \\ 
\hline
\end{tabular}
\caption{The average KL divergence $\pm$ the standard deviation of various algorithms optimized on 30 ground truth networks with 500 samples and random probability distributions for various number of random variables. The best KL scores and the statistically insignificant results are marked in bold.}
\label{tab:stat_test_node}
\vspace*{-5mm}
\end{table}

\subsection{Post-Structure Learning Discretization}
\label{ss:exp_prd}
Figure \ref{fig:exp_samples} and Figure \ref{fig:exp_nodes_random} have shown that DBN-GOMEA-EW and DBN-GOMEA-EF can retrieve accurate networks within relatively short time compared to the other algorithms. DBN-GOMEA-EW and DBN-GOMEA-EF, however, do not optimize the discretization as granularly as e.g., BD. To investigate the effect of doing post-structure learning discretization, all network structures of Figure \ref{fig:exp_samples}, obtained using DBN-GOMEA-EW and DBN-GOMEA-EF on ground truth networks with random probabilities, are once more discretized. The discretizations are optimized with RV-GOMEA and the BD. RV-GOMEA is tasked to optimize the density fitness function (Equation \ref{eq:fitness}) within a budget of 24 hours and is ran on a E5-2690 with 2GB of memory. The BD algorithm is ran on a E5-4650 with 20GB of memory. 

\begin{figure}[h]
  \centering
  \includegraphics[width=\linewidth]{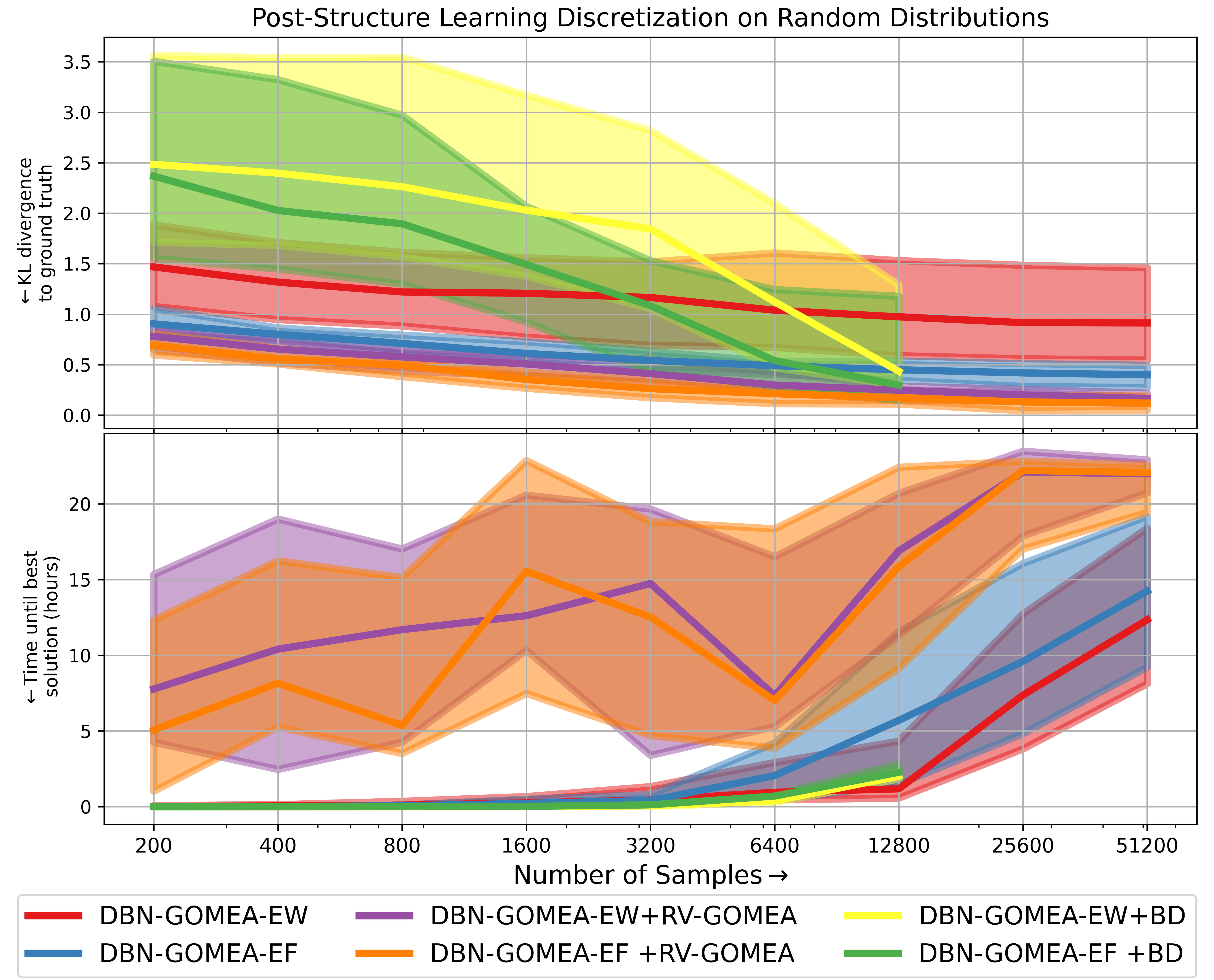}
  \caption{Optimizing the discretization after structure learning. The networks of Figure \ref{fig:exp_samples}, obtained using DBN-GOMEA-EW and DBN-GOMEA-EF, are further optimized. The lines indicate the median, while the shaded regions encompass the first and third quartiles. The arrows on the y-axis point in the direction of improvement per metric.}
  \label{fig:exp_prd}
  \vspace*{-3mm}
\end{figure}

The effect of optimizing the discretization after completing structure learning is shown in Figure \ref{fig:exp_prd}. Figure \ref{fig:exp_prd} also shows the original discretization obtained with DBN-GOMEA-EW and DBN-GOMEA-EF as a reference. Note that the time until the best found solution in Figure \ref{fig:exp_prd} does not include the original 24 hours of structure learning. Figure \ref{fig:exp_prd} shows that when RV-GOMEA is applied (purple and orange), the median KL divergence improves compared to not doing post-structure learning discretization (red and blue) regardless whether the structure was obtained using EW or EF discretization. The BD method however, seems to perform poorly when there are not enough samples in combination with having inaccurate structures (as seen from Figure \ref{fig:exp_samples}). After 12800 samples, BD runs out of memory.

\subsection{Multi-Objective Experiment}
To test the robustness of the MO search, ground truth networks are generated along with expert networks. The expert networks are a simulated representation of what a domain expert believes is the ground truth. The expert networks are randomly generated based on the ground truth networks. In this experiment, the expert networks are configured to know 50\% of the edges of the ground truth network, i.e., 50\% of $l_{\text{edges}}$. This is similar to what has been done in \cite{exploiting_expert_knowledge}. Additional to this, the experts networks are also configured to believe in edges that do not appear in the ground truth network. The number of incorrect edges is also set to 50\% of $l_{\text{edges}}$. The incorrect edges are randomly selected. In this process, networks with cycles are rejected until an acyclic network is found. For the continuous variables, the expert networks also need to determine how the random variables are discretized. For each continuous random variable, the expert network randomly makes between 2 and 4 discretizations. How the data is discretized, i.e., where boundaries are put, is also random. 

For the MO search, MO-DBN-GOMEA with EW and EF discretization are used. In an explainable AI setting, proposed networks that are too complex might be less likely to be accepted by the expert. For this reason, proposed solutions with a complexity (Equation \ref{eq:fitness_penalty}) larger than 10 times the expert network are assigned a constraint value proportionate to the difference in complexity. This threshold however, is problem and expert dependent. In this experiment, it only serves as an example. The number of maximum discretizations is also decreased from 15 to 9, as experts are unlikely to accept complex discretizations. SO algorithms DBN-GOMEA-EW and DBN-GOMEA-EF are also ran for comparison with the same number of maximum discretizations. 

\begin{figure}[h]
  \vspace*{-3mm}
  \centering
  \includegraphics[width=\linewidth]{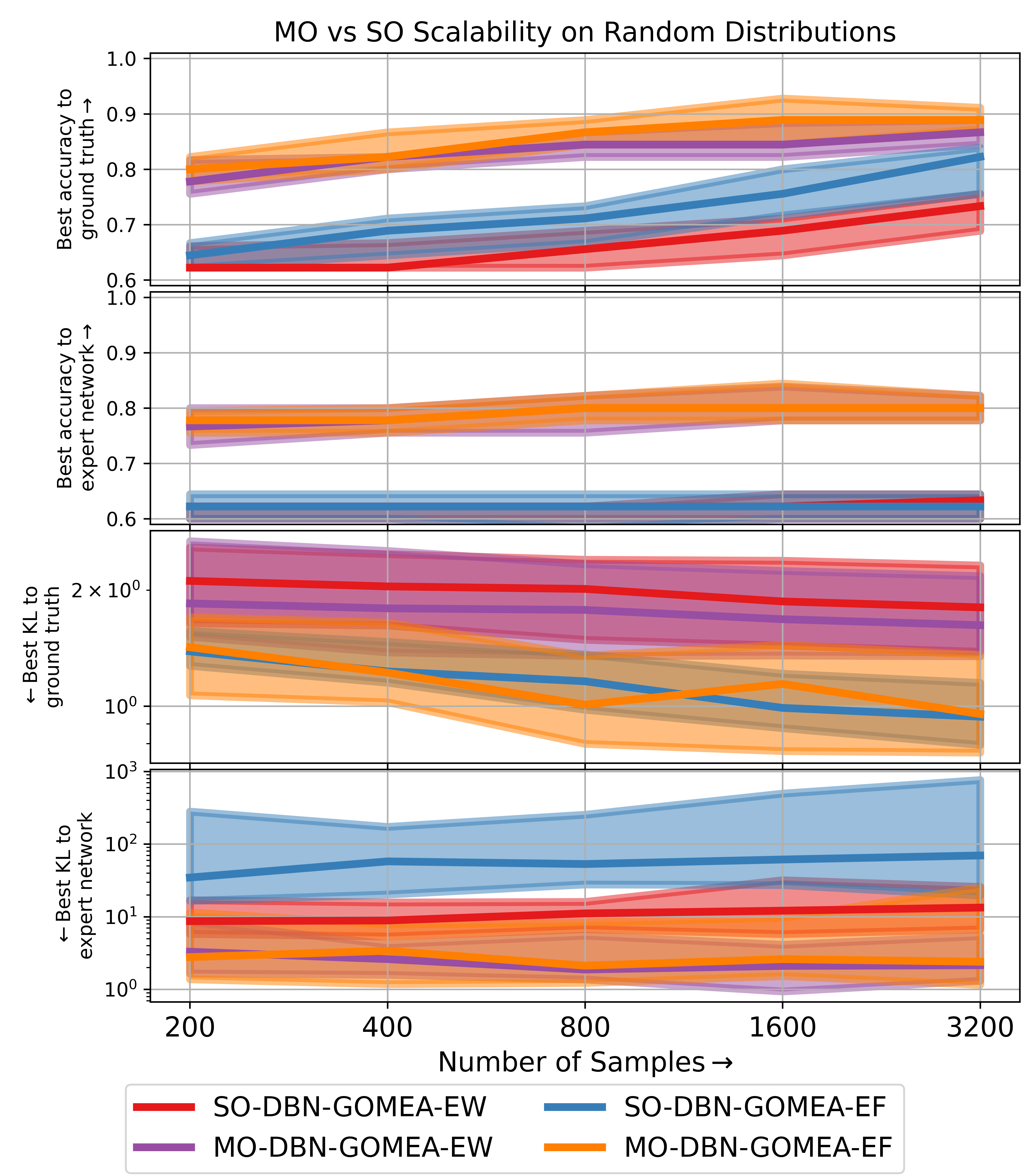}
  \caption{MO vs SO scalability in terms of sample size on ground truth networks with 10 nodes, random probability distributions, and random expert networks. The solid lines are medians, while the shaded areas encompass the first and third quartiles. The arrows on the y-axis point in the direction of improvement per metric.}
  \label{fig:exp_mo}
  \vspace*{-3mm}
\end{figure}

The results of the MO search on 30 randomly generated ground truth networks with 10 random variables and random probability distributions is shown in Figure \ref{fig:exp_mo} for various sample sizes. Each run was performed on a single core of an AMD Genoa 9654, with 2GB of memory and a computation budget of 24 hours. In the top two rows of Figure \ref{fig:exp_mo}, the highest obtained network accuracy is shown with respect to the ground truth and expert networks. For the MO algorithms, the most accurate solutions in the elitist archive are displayed per run. For the SO algorithms, the best found solution's network accuracy is shown. Interestingly, both MO algorithms are able to obtain networks with better accuracy compared to the SO algorithms for the tested sample sizes. In the case of the ground truth network accuracy, the gap between the MO and SO algorithms, does shrink when the sample size grows. In terms of accuracy to the expert network, the MO algorithms perform better than the SO algorithms, as the SO algorithms do not optimize towards the expert network (which is explicitly done in the MO setting).

The best found KL divergence is also shown in Figure \ref{fig:exp_mo} on a log scale. For the MO algorithms, the best KL divergence is shown amongst all solutions with the highest network accuracy, not the entire elitist archive. For the SO algorithms, the best found solution's KL divergence is shown. Interestingly, both the MO and SO algorithms using EF discretization obtain similar KL divergence to the ground truth network. This is not the case for the KL divergence to the expert network as once again, the SO algorithms do not optimize towards the expert network. 

\section{Discussion}
\label{s:discussion}
A state-of-the-art of the art structure learning algorithm based on the Gene-pool Optimal Mixing Evolutionary Algorithm (GOMEA) was extended with discretization-based methods to handle continuous data. In this work, DBN-GOMEA made use of the linkage tree. In the encoding of a solution, the network variables can take three values, namely: $\left \{ 0, 1, 2 \right \}$ . When e.g., Equal Width (EW) or Equal Frequency (EF) discretization is applied, the number of discretizations is also encoded in the solution. The number of discretizations can take a value between 2 and a maximum value for every continuous variable. As the network variables and binning variables have an unequal number of values they can assume, the linkage tree tends to cluster discretization variables together, in the lower parts of the linkage tree. Mixing the network variables and discretization variables could make the optimization even faster, as graphically, discretization variables and edges are structurally related. For this, normalizing the mutual information could help.

In Section \ref{ss:exp_sample_size}, the effect of the sample size on the network accuracy was shown for randomly generated networks with 10 random variables. It was shown that even for large sample sizes, DBN-GOMEA-EF was unable to fully re-obtain the EF ground truth network. Conversely, DBN-GOMEA-BS obtained better KL divergence than DBN-GOMEA-EF for smaller sample sizes. This suggests that more sophisticated discretization techniques, compared to EW and EF discretization, might be required to retrieve the full ground truth network. 

An interesting approach would be to employ a mixed-integer algorithm, such as \cite{gambit}, which handles both integer and real-valued variables. The mixed-integer approach could encode the network structure using integers (as is done in DBN-GOMEA), while encoding the  discretization with real values. 

In this work, a multi-objective Bayesian network learning algorithm was also introduced. Currently, only EW and EF discretization have been ran, as Bayesian discretization is too expensive to run, especially when evaluating complex networks. A multi-objective mixed-integer approach, such as \cite{mo_gambit}, could also be interesting for this problem.

In some real-world domains, blindly trusting machine learning models is not acceptable from a legal aspect. The multi-objective approach proposed in this work however could be useful when used as an advisory model, as it provides the possibility to inspect multiple possible models and trade-off between complexity and accuracy. Exploring the potential added value of our approach from an explainable AI perspective, by having domain experts interact with the found works, is therefore interesting future work.

\section{Conclusion}
In this work, for the first time, a full Bayesian network learning algorithm based GOMEA is presented, which jointly discretizes continuous variables during structure learning. In the single-objective case, the proposed algorithm (DBN-GOMEA) obtains similar or better results than the state-of-the-art when tasked to retrieve randomly generated ground-truth networks. Moreover, leveraging a key strength of EAs, the Bayesian network learning is brought to the multi-objective domain. It was shown how this enables incorporating expert knowledge in a uniquely insightful fashion, finding multiple discrete Bayesian networks that trade-off complexity, accuracy, and the difference with a pre-determined expert network.

\section{Acknowledgments}
This research is part of the research programme Open Competition Domain Science-KLEIN with project number OCENW.KLEIN.111, which is financed by the Dutch Research Council (NWO). Furthermore, we thank NWO for the Small Compute grant on the Dutch National Supercomputer Snellius.


\bibliographystyle{ACM-Reference-Format}
\bibliography{bibliography}


\begin{thebibliography}{28}


\ifx \showCODEN    \undefined \def \showCODEN     #1{\unskip}     \fi
\ifx \showDOI      \undefined \def \showDOI       #1{#1}\fi
\ifx \showISBNx    \undefined \def \showISBNx     #1{\unskip}     \fi
\ifx \showISBNxiii \undefined \def \showISBNxiii  #1{\unskip}     \fi
\ifx \showISSN     \undefined \def \showISSN      #1{\unskip}     \fi
\ifx \showLCCN     \undefined \def \showLCCN      #1{\unskip}     \fi
\ifx \shownote     \undefined \def \shownote      #1{#1}          \fi
\ifx \showarticletitle \undefined \def \showarticletitle #1{#1}   \fi
\ifx \showURL      \undefined \def \showURL       {\relax}        \fi
\providecommand\bibfield[2]{#2}
\providecommand\bibinfo[2]{#2}
\providecommand\natexlab[1]{#1}
\providecommand\showeprint[2][]{arXiv:#2}

\bibitem[\protect\citeauthoryear{Amirkhani, Rahmati, Lucas, and Hommersom}{Amirkhani et~al\mbox{.}}{2017}]%
        {exploiting_expert_knowledge}
\bibfield{author}{\bibinfo{person}{Hossein Amirkhani}, \bibinfo{person}{Mohammad Rahmati}, \bibinfo{person}{Peter J.~F. Lucas}, {and} \bibinfo{person}{Arjen Hommersom}.} \bibinfo{year}{2017}\natexlab{}.
\newblock \showarticletitle{Exploiting Experts’ Knowledge for Structure Learning of Bayesian Networks}.
\newblock \bibinfo{journal}{\emph{IEEE Transactions on Pattern Analysis and Machine Intelligence}} \bibinfo{volume}{39}, \bibinfo{number}{11} (\bibinfo{year}{2017}), \bibinfo{pages}{2154--2170}.
\newblock
\urldef\tempurl%
\url{https://doi.org/10.1109/TPAMI.2016.2636828}
\showDOI{\tempurl}


\bibitem[\protect\citeauthoryear{Beuzen, Marshall, and Splinter}{Beuzen et~al\mbox{.}}{2018}]%
        {example_coastal_erosion}
\bibfield{author}{\bibinfo{person}{Tomas Beuzen}, \bibinfo{person}{Lucy Marshall}, {and} \bibinfo{person}{Kristen~D. Splinter}.} \bibinfo{year}{2018}\natexlab{}.
\newblock \showarticletitle{A comparison of methods for discretizing continuous variables in Bayesian Networks}.
\newblock \bibinfo{journal}{\emph{Environmental Modelling \& Software}}  \bibinfo{volume}{108} (\bibinfo{year}{2018}), \bibinfo{pages}{61--66}.
\newblock
\showISSN{1364-8152}
\urldef\tempurl%
\url{https://doi.org/10.1016/j.envsoft.2018.07.007}
\showDOI{\tempurl}


\bibitem[\protect\citeauthoryear{Bouter, Alderliesten, Witteveen, and Bosman}{Bouter et~al\mbox{.}}{2017}]%
        {rv_gomea}
\bibfield{author}{\bibinfo{person}{Anton Bouter}, \bibinfo{person}{Tanja Alderliesten}, \bibinfo{person}{Cees Witteveen}, {and} \bibinfo{person}{Peter A.~N. Bosman}.} \bibinfo{year}{2017}\natexlab{}.
\newblock \showarticletitle{Exploiting Linkage Information in Real-Valued Optimization with the Real-Valued Gene-Pool Optimal Mixing Evolutionary Algorithm}. In \bibinfo{booktitle}{\emph{Proceedings of the Genetic and Evolutionary Computation Conference}} (Berlin, Germany) \emph{(\bibinfo{series}{GECCO '17})}. \bibinfo{publisher}{Association for Computing Machinery}, \bibinfo{address}{New York, NY, USA}, \bibinfo{pages}{705–712}.
\newblock
\showISBNx{9781450349208}
\urldef\tempurl%
\url{https://doi.org/10.1145/3071178.3071272}
\showDOI{\tempurl}


\bibitem[\protect\citeauthoryear{Bouter and Bosman}{Bouter and Bosman}{2023}]%
        {gomea_library}
\bibfield{author}{\bibinfo{person}{Anton Bouter} {and} \bibinfo{person}{Peter A.~N. Bosman}.} \bibinfo{year}{2023}\natexlab{}.
\newblock \showarticletitle{A Joint Python/C++ Library for Efficient yet Accessible Black-Box and Gray-Box Optimization with GOMEA}. In \bibinfo{booktitle}{\emph{Proceedings of the Companion Conference on Genetic and Evolutionary Computation}} (Lisbon, Portugal) \emph{(\bibinfo{series}{GECCO '23 Companion})}. \bibinfo{publisher}{Association for Computing Machinery}, \bibinfo{address}{New York, NY, USA}, \bibinfo{pages}{1864–1872}.
\newblock
\showISBNx{9798400701207}
\urldef\tempurl%
\url{https://doi.org/10.1145/3583133.3596361}
\showDOI{\tempurl}


\bibitem[\protect\citeauthoryear{Bubnova, Deeva, and Kalyuzhnaya}{Bubnova et~al\mbox{.}}{2021}]%
        {mixbn}
\bibfield{author}{\bibinfo{person}{Anna~V. Bubnova}, \bibinfo{person}{Irina Deeva}, {and} \bibinfo{person}{Anna~V. Kalyuzhnaya}.} \bibinfo{year}{2021}\natexlab{}.
\newblock \showarticletitle{MIxBN: library for learning Bayesian networks from mixed data}.
\newblock \bibinfo{journal}{\emph{Procedia Computer Science}}  \bibinfo{volume}{193} (\bibinfo{year}{2021}), \bibinfo{pages}{494--503}.
\newblock
\showISSN{1877-0509}
\urldef\tempurl%
\url{https://doi.org/10.1016/j.procs.2021.10.051}
\showDOI{\tempurl}
\newblock
\shownote{10th International Young Scientists Conference in Computational Science, YSC2021, 28 June – 2 July, 2021}.


\bibitem[\protect\citeauthoryear{Chen, Wheeler, and Kochenderfer}{Chen et~al\mbox{.}}{2017}]%
        {learn_dbn}
\bibfield{author}{\bibinfo{person}{Yi-Chun Chen}, \bibinfo{person}{Tim~A. Wheeler}, {and} \bibinfo{person}{Mykel~J. Kochenderfer}.} \bibinfo{year}{2017}\natexlab{}.
\newblock \showarticletitle{Learning Discrete Bayesian Networks from Continuous Data}.
\newblock \bibinfo{journal}{\emph{J. Artif. Int. Res.}} \bibinfo{volume}{59}, \bibinfo{number}{1} (\bibinfo{year}{2017}), \bibinfo{pages}{103–132}.
\newblock
\showISSN{1076-9757}


\bibitem[\protect\citeauthoryear{de~Campos, Cano, Castellano, and Moral}{de~Campos et~al\mbox{.}}{2011}]%
        {example_gene_expression}
\bibfield{author}{\bibinfo{person}{Luis~M. de Campos}, \bibinfo{person}{Andrés Cano}, \bibinfo{person}{Javier~G. Castellano}, {and} \bibinfo{person}{Serafín Moral}.} \bibinfo{year}{2011}\natexlab{}.
\newblock \showarticletitle{Bayesian networks classifiers for gene-expression data}, In \bibinfo{booktitle}{2011 11th International Conference on Intelligent Systems Design and Applications}.
\newblock \bibinfo{journal}{\emph{International Conference on Intelligent Systems Design and Applications, ISDA}}, \bibinfo{pages}{1200--1206}.
\newblock
\showISSN{2164-7151}
\urldef\tempurl%
\url{https://doi.org/10.1109/ISDA.2011.6121822}
\showDOI{\tempurl}


\bibitem[\protect\citeauthoryear{Fayyad and Irani}{Fayyad and Irani}{1993}]%
        {fayyad_irani}
\bibfield{author}{\bibinfo{person}{Usama~M Fayyad} {and} \bibinfo{person}{Keki~B Irani}.} \bibinfo{year}{1993}\natexlab{}.
\newblock \showarticletitle{Multi-interval discretization of continuous-valued attributes for classification learning}. In \bibinfo{booktitle}{\emph{Proceedings of the 13th International Joint Conference on Artificial Intelligence}}. \bibinfo{publisher}{Morgan Kaufmann}, \bibinfo{address}{Chambery, France}, \bibinfo{pages}{1022--1029}.
\newblock


\bibitem[\protect\citeauthoryear{Friedman and Goldszmidt}{Friedman and Goldszmidt}{1996}]%
        {discretizing_friedman}
\bibfield{author}{\bibinfo{person}{Nir Friedman} {and} \bibinfo{person}{Moises Goldszmidt}.} \bibinfo{year}{1996}\natexlab{}.
\newblock \showarticletitle{Discretizing Continuous Attributes While Learning Bayesian Networks}. In \bibinfo{booktitle}{\emph{Proceedings of the Thirteenth International Conference on Machine Learning}}, \bibfield{editor}{\bibinfo{person}{Lorenza Saitta}} (Ed.). \bibinfo{publisher}{Morgan Kaufmann}, \bibinfo{address}{San Francisco, CA}.
\newblock


\bibitem[\protect\citeauthoryear{Hosseini and Ivanov}{Hosseini and Ivanov}{2020}]%
        {example_supply_chain}
\bibfield{author}{\bibinfo{person}{Seyedmohsen Hosseini} {and} \bibinfo{person}{Dmitry Ivanov}.} \bibinfo{year}{2020}\natexlab{}.
\newblock \showarticletitle{Bayesian networks for supply chain risk, resilience and ripple effect analysis: A literature review}.
\newblock \bibinfo{journal}{\emph{Expert Systems with Applications}}  \bibinfo{volume}{161} (\bibinfo{year}{2020}), \bibinfo{pages}{113649}.
\newblock
\showISSN{0957-4174}
\urldef\tempurl%
\url{https://doi.org/10.1016/j.eswa.2020.113649}
\showDOI{\tempurl}


\bibitem[\protect\citeauthoryear{Ickstadt, Bornkamp, Grzegorczyk, Wieczorek, Sheriff, Grecco, and Zamir}{Ickstadt et~al\mbox{.}}{2011}]%
        {nonparametric_bayesian_networks}
\bibfield{author}{\bibinfo{person}{Katja Ickstadt}, \bibinfo{person}{Bjöorn Bornkamp}, \bibinfo{person}{Marco Grzegorczyk}, \bibinfo{person}{Jakob Wieczorek}, \bibinfo{person}{Malik~R. Sheriff}, \bibinfo{person}{Hernáan~E. Grecco}, {and} \bibinfo{person}{Eli Zamir}.} \bibinfo{year}{2011}\natexlab{}.
\newblock \showarticletitle{{Nonparametric Bayesian Networks}}.
\newblock In \bibinfo{booktitle}{\emph{{Bayesian Statistics 9}}}. \bibinfo{publisher}{Oxford University Press}, \bibinfo{address}{Oxford, United Kingdom}.
\newblock
\showISBNx{9780199694587}
\urldef\tempurl%
\url{https://doi.org/10.1093/acprof:oso/9780199694587.003.0010}
\showDOI{\tempurl}
\showeprint{https://academic.oup.com/book/0/chapter/141642184/chapter-ag-pdf/45787762/book\_1879\_section\_141642184.ag.pdf}


\bibitem[\protect\citeauthoryear{Ide, Cozman, and Ramos}{Ide et~al\mbox{.}}{2004}]%
        {random_bn_structure}
\bibfield{author}{\bibinfo{person}{Jaime~S. Ide}, \bibinfo{person}{Fabio~G. Cozman}, {and} \bibinfo{person}{Fabio~T. Ramos}.} \bibinfo{year}{2004}\natexlab{}.
\newblock \showarticletitle{Generating Random Bayesian networks with constraints on induced width}. In \bibinfo{booktitle}{\emph{Proceedings of the 16th European Conference on Artificial Intelligence}} (Valencia, Spain) \emph{(\bibinfo{series}{ECAI'04})}. \bibinfo{publisher}{IOS Press}, \bibinfo{address}{NLD}, \bibinfo{pages}{353–357}.
\newblock
\showISBNx{9781586034528}


\bibitem[\protect\citeauthoryear{Koller and Friedman}{Koller and Friedman}{2009}]%
        {koller_friedman_probabilistic_graphical_models}
\bibfield{author}{\bibinfo{person}{Daphne Koller} {and} \bibinfo{person}{Nir Friedman}.} \bibinfo{year}{2009}\natexlab{}.
\newblock \bibinfo{booktitle}{\emph{Probabilistic Graphical Models: Principles and Techniques - Adaptive Computation and Machine Learning}}.
\newblock \bibinfo{publisher}{The MIT Press}, \bibinfo{address}{Cambridge, Massachusetts}.
\newblock
\showISBNx{0262013193}


\bibitem[\protect\citeauthoryear{Lima, Nassar, Rodrigues, Filho, and Jacinto}{Lima et~al\mbox{.}}{2014}]%
        {example_bit_rate}
\bibfield{author}{\bibinfo{person}{Mariana~D.C. Lima}, \bibinfo{person}{Silvia~M. Nassar}, \bibinfo{person}{Pedro Ivo~R.B.G. Rodrigues}, \bibinfo{person}{Paulo J.~Freitas Filho}, {and} \bibinfo{person}{Carlos~M.C. Jacinto}.} \bibinfo{year}{2014}\natexlab{}.
\newblock \showarticletitle{Heuristic Discretization Method for Bayesian Networks}.
\newblock \bibinfo{journal}{\emph{Journal of Computer Science}} \bibinfo{volume}{10}, \bibinfo{number}{5} (\bibinfo{year}{2014}), \bibinfo{pages}{869--878}.
\newblock
\urldef\tempurl%
\url{https://doi.org/10.3844/jcssp.2014.869.878}
\showDOI{\tempurl}


\bibitem[\protect\citeauthoryear{Liu, Malone, and Yuan}{Liu et~al\mbox{.}}{2012}]%
        {empirical_score_functions}
\bibfield{author}{\bibinfo{person}{Zhifa Liu}, \bibinfo{person}{Brandon Malone}, {and} \bibinfo{person}{Changhe Yuan}.} \bibinfo{year}{2012}\natexlab{}.
\newblock \showarticletitle{Empirical Evaluation of Scoring Functions for Bayesian Network Model Selection}.
\newblock \bibinfo{journal}{\emph{BMC Bioinformatics}}  \bibinfo{volume}{13} (\bibinfo{year}{2012}), \bibinfo{pages}{S14}.
\newblock
\urldef\tempurl%
\url{https://doi.org/10.1186/1471-2105-13-S15-S14}
\showDOI{\tempurl}


\bibitem[\protect\citeauthoryear{Luong, {La Poutré}, and Bosman}{Luong et~al\mbox{.}}{2018}]%
        {mo_gomea}
\bibfield{author}{\bibinfo{person}{Ngoc~Hoang Luong}, \bibinfo{person}{Han {La Poutré}}, {and} \bibinfo{person}{Peter~A.N. Bosman}.} \bibinfo{year}{2018}\natexlab{}.
\newblock \showarticletitle{Multi-objective Gene-pool Optimal Mixing Evolutionary Algorithm with the Interleaved Multi-start Scheme}.
\newblock \bibinfo{journal}{\emph{Swarm and Evolutionary Computation}}  \bibinfo{volume}{40} (\bibinfo{year}{2018}), \bibinfo{pages}{238--254}.
\newblock
\showISSN{2210-6502}
\urldef\tempurl%
\url{https://doi.org/10.1016/j.swevo.2018.02.005}
\showDOI{\tempurl}


\bibitem[\protect\citeauthoryear{Orphanou, Thierens, and Bosman}{Orphanou et~al\mbox{.}}{2018}]%
        {bn_gomea}
\bibfield{author}{\bibinfo{person}{Kalia Orphanou}, \bibinfo{person}{Dirk Thierens}, {and} \bibinfo{person}{Peter A.~N. Bosman}.} \bibinfo{year}{2018}\natexlab{}.
\newblock \showarticletitle{Learning Bayesian Network Structures with GOMEA}. In \bibinfo{booktitle}{\emph{Proceedings of the Genetic and Evolutionary Computation Conference}} (Kyoto, Japan) \emph{(\bibinfo{series}{GECCO '18})}. \bibinfo{publisher}{Association for Computing Machinery}, \bibinfo{address}{New York, NY, USA}, \bibinfo{pages}{1007–1014}.
\newblock
\showISBNx{9781450356183}
\urldef\tempurl%
\url{https://doi.org/10.1145/3205455.3205502}
\showDOI{\tempurl}


\bibitem[\protect\citeauthoryear{Pearl}{Pearl}{1988}]%
        {pearl_probabilistic_reasoning}
\bibfield{author}{\bibinfo{person}{Judea Pearl}.} \bibinfo{year}{1988}\natexlab{}.
\newblock \bibinfo{booktitle}{\emph{Probabilistic Reasoning in Intelligent Systems: Networks of Plausible Inference}}.
\newblock \bibinfo{publisher}{Morgan Kaufmann Publishers Inc.}, \bibinfo{address}{San Francisco, CA, USA}.
\newblock
\showISBNx{1558604790}


\bibitem[\protect\citeauthoryear{Reijnen, Gogou, Visser, Engerud, Ramjith, van~der Putten, van~de Vijver, Santacana, Bronsert, Bulten, Hirschfeld, Colas, Gil-Moreno, Reques, Mancebo, Krakstad, Trovik, Haldorsen, Huvila, Koskas, Weinberger, Bednarikova, Hausnerova, van~der Wurff, Matias-Guiu, Amant, Consortium, Massuger, Snijders, Küsters-Vandevelde, Lucas, and Pijnenborg}{Reijnen et~al\mbox{.}}{2020}]%
        {example_endorisk}
\bibfield{author}{\bibinfo{person}{Casper Reijnen}, \bibinfo{person}{Evangelia Gogou}, \bibinfo{person}{Nicole C.~M. Visser}, \bibinfo{person}{Hilde Engerud}, \bibinfo{person}{Jordache Ramjith}, \bibinfo{person}{Louis J.~M. van~der Putten}, \bibinfo{person}{Koen van~de Vijver}, \bibinfo{person}{Maria Santacana}, \bibinfo{person}{Peter Bronsert}, \bibinfo{person}{Johan Bulten}, \bibinfo{person}{Marc Hirschfeld}, \bibinfo{person}{Eva Colas}, \bibinfo{person}{Antonio Gil-Moreno}, \bibinfo{person}{Armando Reques}, \bibinfo{person}{Gemma Mancebo}, \bibinfo{person}{Camilla Krakstad}, \bibinfo{person}{Jone Trovik}, \bibinfo{person}{Ingfrid~S. Haldorsen}, \bibinfo{person}{Jutta Huvila}, \bibinfo{person}{Martin Koskas}, \bibinfo{person}{Vit Weinberger}, \bibinfo{person}{Marketa Bednarikova}, \bibinfo{person}{Jitka Hausnerova}, \bibinfo{person}{Anneke A.~M. van~der Wurff}, \bibinfo{person}{Xavier Matias-Guiu}, \bibinfo{person}{Frederic Amant}, \bibinfo{person}{ENITEC Consortium}, \bibinfo{person}{Leon F. A.~G.
  Massuger}, \bibinfo{person}{Marc P. L.~M. Snijders}, \bibinfo{person}{Heidi V.~N. Küsters-Vandevelde}, \bibinfo{person}{Peter J.~F. Lucas}, {and} \bibinfo{person}{Johanna M.~A. Pijnenborg}.} \bibinfo{year}{2020}\natexlab{}.
\newblock \showarticletitle{Preoperative risk stratification in endometrial cancer (ENDORISK) by a Bayesian network model: A development and validation study}.
\newblock \bibinfo{journal}{\emph{PLOS Medicine}} \bibinfo{volume}{17}, \bibinfo{number}{5} (\bibinfo{date}{05} \bibinfo{year}{2020}), \bibinfo{pages}{1--19}.
\newblock
\urldef\tempurl%
\url{https://doi.org/10.1371/journal.pmed.1003111}
\showDOI{\tempurl}


\bibitem[\protect\citeauthoryear{Ropero, Renooij, and {van der Gaag}}{Ropero et~al\mbox{.}}{2018}]%
        {example_discretizing_environmental_data}
\bibfield{author}{\bibinfo{person}{Rosa~F. Ropero}, \bibinfo{person}{Silja Renooij}, {and} \bibinfo{person}{Linda~C. {van der Gaag}}.} \bibinfo{year}{2018}\natexlab{}.
\newblock \showarticletitle{Discretizing environmental data for learning Bayesian-network classifiers}.
\newblock \bibinfo{journal}{\emph{Ecological Modelling}}  \bibinfo{volume}{368} (\bibinfo{year}{2018}), \bibinfo{pages}{391--403}.
\newblock
\showISSN{0304-3800}
\urldef\tempurl%
\url{https://doi.org/10.1016/j.ecolmodel.2017.12.015}
\showDOI{\tempurl}


\bibitem[\protect\citeauthoryear{Rostamabadi, Jahangiri, Zarei, Kamalinia, and Alimohammadlou}{Rostamabadi et~al\mbox{.}}{2020}]%
        {example_chemical_process}
\bibfield{author}{\bibinfo{person}{Akbar Rostamabadi}, \bibinfo{person}{Mehdi Jahangiri}, \bibinfo{person}{Esmaeil Zarei}, \bibinfo{person}{Mojtaba Kamalinia}, {and} \bibinfo{person}{Moslem Alimohammadlou}.} \bibinfo{year}{2020}\natexlab{}.
\newblock \showarticletitle{A novel Fuzzy Bayesian Network approach for safety analysis of process systems; An application of HFACS and SHIPP methodology}.
\newblock \bibinfo{journal}{\emph{Journal of Cleaner Production}}  \bibinfo{volume}{244} (\bibinfo{year}{2020}), \bibinfo{pages}{118761}.
\newblock
\showISSN{0959-6526}
\urldef\tempurl%
\url{https://doi.org/10.1016/j.jclepro.2019.118761}
\showDOI{\tempurl}


\bibitem[\protect\citeauthoryear{Sadowski, Thierens, and Bosman}{Sadowski et~al\mbox{.}}{2018}]%
        {gambit}
\bibfield{author}{\bibinfo{person}{Krzysztof~L. Sadowski}, \bibinfo{person}{Dirk Thierens}, {and} \bibinfo{person}{Peter~A.N. Bosman}.} \bibinfo{year}{2018}\natexlab{}.
\newblock \showarticletitle{{GAMBIT: A Parameterless Model-Based Evolutionary Algorithm for Mixed-Integer Problems}}.
\newblock \bibinfo{journal}{\emph{Evolutionary Computation}} \bibinfo{volume}{26}, \bibinfo{number}{1} (\bibinfo{date}{03} \bibinfo{year}{2018}), \bibinfo{pages}{117--143}.
\newblock
\showISSN{1063-6560}
\urldef\tempurl%
\url{https://doi.org/10.1162/evco_a_00206}
\showDOI{\tempurl}
\showeprint{https://direct.mit.edu/evco/article-pdf/26/1/117/1547009/evco\_a\_00206.pdf}


\bibitem[\protect\citeauthoryear{Sadowski, Thierens, and Bosman}{Sadowski et~al\mbox{.}}{2021}]%
        {mo_gambit}
\bibfield{author}{\bibinfo{person}{Krzysztof~L. Sadowski}, \bibinfo{person}{Dirk Thierens}, {and} \bibinfo{person}{Peter A.~N. Bosman}.} \bibinfo{year}{2021}\natexlab{}.
\newblock \showarticletitle{Optimization of multi-objective mixed-integer problems with a model-based evolutionary algorithm in a black-box setting}. In \bibinfo{booktitle}{\emph{Proceedings of the Genetic and Evolutionary Computation Conference Companion}} (Lille, France) \emph{(\bibinfo{series}{GECCO '21})}. \bibinfo{publisher}{Association for Computing Machinery}, \bibinfo{address}{New York, NY, USA}, \bibinfo{pages}{227–228}.
\newblock
\showISBNx{9781450383516}
\urldef\tempurl%
\url{https://doi.org/10.1145/3449726.3459521}
\showDOI{\tempurl}


\bibitem[\protect\citeauthoryear{Schwarz}{Schwarz}{1978}]%
        {bic}
\bibfield{author}{\bibinfo{person}{Gideon Schwarz}.} \bibinfo{year}{1978}\natexlab{}.
\newblock \showarticletitle{Estimating the dimension of a model}.
\newblock \bibinfo{journal}{\emph{The annals of statistics}} \bibinfo{volume}{6}, \bibinfo{number}{2} (\bibinfo{year}{1978}), \bibinfo{pages}{461--464}.
\newblock


\bibitem[\protect\citeauthoryear{Suzuki}{Suzuki}{2014}]%
        {joes_cousin}
\bibfield{author}{\bibinfo{person}{Joe Suzuki}.} \bibinfo{year}{2014}\natexlab{}.
\newblock \showarticletitle{Learning Bayesian Network Structures When Discrete and Continuous Variables Are Present}. In \bibinfo{booktitle}{\emph{Probabilistic Graphical Models}}, \bibfield{editor}{\bibinfo{person}{Linda~C. van~der Gaag} {and} \bibinfo{person}{Ad~J. Feelders}} (Eds.). \bibinfo{publisher}{Springer International Publishing}, \bibinfo{address}{Cham}, \bibinfo{pages}{471--486}.
\newblock
\showISBNx{978-3-319-11433-0}


\bibitem[\protect\citeauthoryear{Wu, Qian, Liu, Zhou, and Zhou}{Wu et~al\mbox{.}}{2023}]%
        {mo_bi_objective_scalable}
\bibfield{author}{\bibinfo{person}{Ting Wu}, \bibinfo{person}{Hong Qian}, \bibinfo{person}{Ziqi Liu}, \bibinfo{person}{Jun Zhou}, {and} \bibinfo{person}{Aimin Zhou}.} \bibinfo{year}{2023}\natexlab{}.
\newblock \showarticletitle{Bi-Objective Evolutionary Bayesian Network Structure Learning via Skeleton Constraint}.
\newblock \bibinfo{journal}{\emph{Front. Comput. Sci.}} \bibinfo{volume}{17}, \bibinfo{number}{6} (\bibinfo{year}{2023}), \bibinfo{numpages}{13}~pages.
\newblock
\showISSN{2095-2228}
\urldef\tempurl%
\url{https://doi.org/10.1007/s11704-023-2740-6}
\showDOI{\tempurl}


\bibitem[\protect\citeauthoryear{Zhao, Feng, Chen, Zhou, and Yu}{Zhao et~al\mbox{.}}{2022}]%
        {example_radiologist}
\bibfield{author}{\bibinfo{person}{G. Zhao}, \bibinfo{person}{Q. Feng}, \bibinfo{person}{C. Chen}, \bibinfo{person}{Z. Zhou}, {and} \bibinfo{person}{Y. Yu}.} \bibinfo{year}{2022}\natexlab{}.
\newblock \showarticletitle{Diagnose Like a Radiologist: Hybrid Neuro-Probabilistic Reasoning for Attribute-Based Medical Image Diagnosis}.
\newblock \bibinfo{journal}{\emph{IEEE Transactions on Pattern Analysis \&; Machine Intelligence}} \bibinfo{volume}{44}, \bibinfo{number}{11} (\bibinfo{year}{2022}), \bibinfo{pages}{7400--7416}.
\newblock
\showISSN{1939-3539}
\urldef\tempurl%
\url{https://doi.org/10.1109/TPAMI.2021.3130759}
\showDOI{\tempurl}


\bibitem[\protect\citeauthoryear{Zhou, Yu, Zhu, Zhou, and Qi}{Zhou et~al\mbox{.}}{2023}]%
        {example_aviation}
\bibfield{author}{\bibinfo{person}{Zhipeng Zhou}, \bibinfo{person}{Xinhui Yu}, \bibinfo{person}{Zeyu Zhu}, \bibinfo{person}{Dequn Zhou}, {and} \bibinfo{person}{Haonan Qi}.} \bibinfo{year}{2023}\natexlab{}.
\newblock \showarticletitle{Development and application of a Bayesian network-based model for systematically reducing safety risks in the commercial air transportation system}.
\newblock \bibinfo{journal}{\emph{Safety Science}}  \bibinfo{volume}{157} (\bibinfo{year}{2023}), \bibinfo{pages}{105942}.
\newblock
\showISSN{0925-7535}
\urldef\tempurl%
\url{https://doi.org/10.1016/j.ssci.2022.105942}
\showDOI{\tempurl}


\end{thebibliography}

\appendix









\end{document}